\newif\ifpnas
\pnasfalse

\documentclass[twocolumn,reprint,superscriptaddress,showpacs,showkeys]{revtex4-1}
\usepackage[utf8]{inputenc}
\usepackage{newtxtext}
\RequirePackage[dvipsnames,usenames]{xcolor}
\usepackage[hidelinks]{hyperref}

\usepackage{color} 
\usepackage{amsmath,amssymb} 
\usepackage{graphicx} 
\usepackage[super]{nth}
\usepackage{cleveref}

\usepackage[english]{babel}

\newcommand{\nMSE}{\mathrm{nMSE}}
\newcommand{\MSE}{\mathrm{MSE}}
\newcommand{\Roff}{R_\mathrm{off}}
\newcommand{\Ron}{R_\mathrm{on}}
\newcommand{\pvec}[1]{\vec{#1}\mkern2mu\vphantom{#1}}

\def\memstate{\eta}
\def\Memstate{H}

\begin{document}

\title{The Computational Capacity of LRC, Memristive and Hybrid Reservoirs}

\author{Forrest C. Sheldon}
\email[To whom correspondence should be addressed. E-mail: ]{fs@lims.ac.uk}
\altaffiliation[Current Address: ]{London Institute for Mathematical Sciences, 21 Albemarle St. London, W1S 4BS, UK}
\affiliation{Theoretical Division and Center for Nonlinear Studies, Los Alamos National Laboratory, Los Alamos, New Mexico 87545, USA}

\author{Artemy Kolchinsky}
\affiliation{Santa Fe Institute, 1399 Hyde Park Road, Santa Fe, NM 87501, USA}

\author{Francesco Caravelli}
\affiliation{Theoretical Division and Center for Nonlinear Studies, Los Alamos National Laboratory, Los Alamos, New Mexico 87545, USA}


\begin{abstract}

Reservoir computing is a machine learning paradigm that uses a high-dimensional dynamical system, or \emph{reservoir}, to approximate and predict time series data.
The scale, speed and power usage  of 
reservoir computers could be enhanced by constructing reservoirs out of electronic circuits, and several experimental studies have demonstrated promise in this direction. However, designing quality reservoirs requires a
precise understanding of how such circuits process and store information. We analyze the feasibility and optimal design of electronic reservoirs 
that include both linear elements (resistors, inductors, and capacitors) and nonlinear memory elements called memristors. 
We provide analytic results regarding the feasibility of these reservoirs, 
and give a systematic characterization of their computational properties by
examining the types of input-output relationships that they can approximate.  This allows us to design reservoirs with optimal properties. By introducing measures of the total linear and nonlinear computational capacities of the reservoir, we are able to design electronic circuits whose total computational capacity scales extensively with the system size. Our electronic reservoirs can match or exceed the performance of conventional ``echo state network'' reservoirs in a form that may be directly implemented in hardware.
\end{abstract}









  \maketitle

  \section{Introduction}


 Reservoir computing (RC) \cite{jaeger2001echo,jaeger2002short,lukovsevivcius2012practical} is a model for performing computations on time series data, which combines a  high-dimensional driven dynamical system, called a \emph{reservoir}, with a simple learning algorithm. 
Reservoir computing has proven to be a powerful tool in a wide variety of signal processing tasks, including forecasting \cite{jaeger2001echo}, pattern generation and classification \cite{bertschinger2004real}, adaptive filtering and prediction of chaotic systems \cite{jaeger2004harnessing}.  Recently, extensions to spatio-temporal chaotic systems \cite{pathak2018model} have proven to be surprisingly effective.

Central to the success of reservoir computation is the use of large dynamical systems to generate nonlinear transformations and store memories of the driving signal. This has generated interest in developing nanoscale electronic reservoirs with large numbers of elements \cite{tanaka2019recent}, incorporating both linear components (such as resistors, inductors, and capacitors) and nonlinear components such as \emph{memristors}. 
Memristors, or ``{resistors with memory}'', are nanoscale devices whose resistance depends on the past history of the current.  The currents flowing through these devices cause a rearrangement of ions, leading to a persistent (but also reversible) change in resistance.
Memristors offer the possibility of harnessing both nonlinear behavior and memory in electronic circuits. For this reason, specialized circuits composed of large numbers of memristors promise a new generation of computational hardware operating orders of magnitude faster, and at far lower power, than traditional digital circuitry \cite{chua1971memristor,strukov2008missing,Review1,Review2,singlerc,singlercm}. 

Recently, several experimental works examining memristor-based electronic reservoirs  have shown remarkable promise \cite{marinella2019efficient,kulkarni2012memristor,du2017reservoir,Moon2019,Carbajal2015,konkoli2020,inubushi2017reservoir}. However, there is still no fundamental understanding of when electronic circuits can be successfully employed as reservoirs, what type of functions they can compute, and how to design electronic reservoirs with specific computational properties. In this paper we address this gap by providing a systematic and analytical study of the computational capacity of electronic reservoirs composed of traditional linear elements and memristors.

In order to be feasible as a reservoir, a driven dynamical system must satisfy certain properties which guarantee that its state encodes an informative function of the driving signal; we establish feasibility for models of linear/LRC (inductor-resistor-capacitor)
and memristor reservoirs.
We also characterize the input-output relationships natural to electronic reservoirs, in the process showing how memristors may be viewed as a source of nonlinear computations.  We then demonstrate how to combine linear and nonlinear elements to achieve a specific computational task (that of approximating a 2nd order filter).  Lastly, the capacity of the reservoir to perform useful computations should increase as the size (i.e., dimensionality) of the reservoir is increased.  
In particular, the main motivation for electronic reservoirs is the potential to achieve very large reservoir sizes, but increases in size are useless if they do not lead to improved computational capacities of the reservoir.  
Optimally, the number of linearly independent inputs that the reservoir can reconstruct should scale linearly with the reservoir size --- i.e., to borrow a term from statistical physics, their reconstruction capabilities should be {extensive} in reservoir size. For both LRC and memristor reservoirs, we consider measures of linear and nonlinear computational capacity, and show that they can be made to scale extensively. The approach we use to analyze the computational capacities of memristor and LRC reservoirs may be generalized to other reservoirs and computational tasks in analogy to the methods available to tune echo state networks (ESNs) by trading between memory storage and nonlinearity.

In the next section we make these notions more precise. We provide an introduction to reservoir computing, electronic reservoirs and memristors, as well as definitions of the measures of computational capacity we will utilize. We then turn to the feasibility, tunability and scalability of LRC and memristor-based reservoirs. Finally, we compare our results to more conventional ESN reservoirs \cite{jaeger2001echo}, showing that electronic reservoirs are capable of matching or exceeding the performance of a standard ESN reservoir implementations.

\begin{figure}
\centering
\includegraphics[width=\columnwidth]{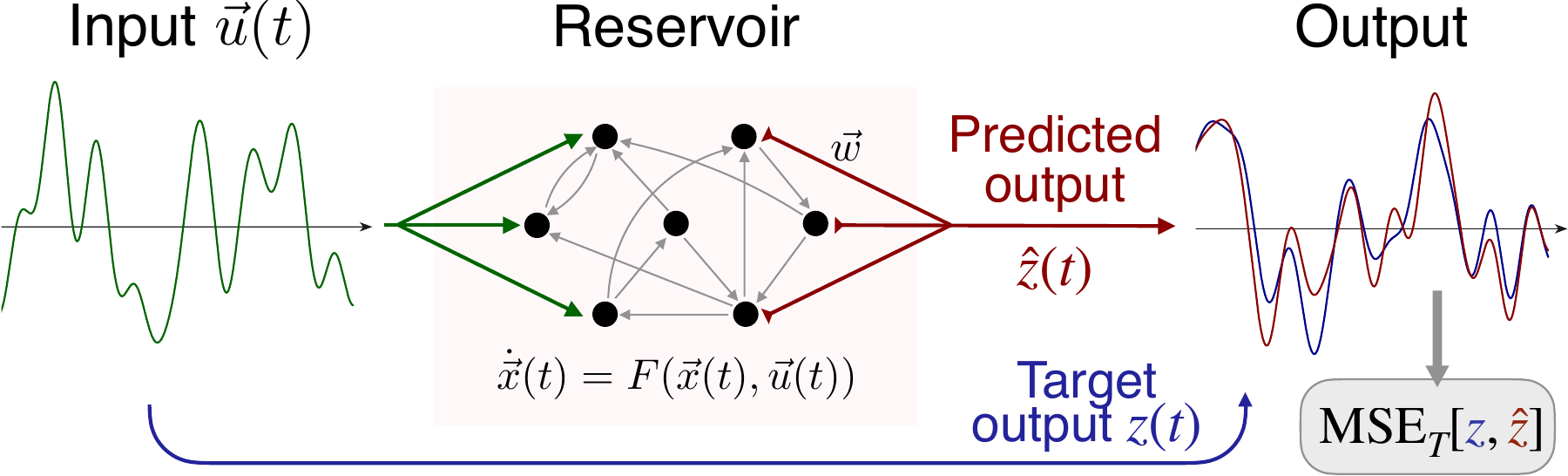}
\caption{Basics of reservoir computing. A time-dependent input $\vec{u}(t)$ is used to drive a reservoir, which is a (linear or nonlinear) dynamical system with internal state $\vec{x}(t)$. The predicted output is computed as a weighted linear combination of the internal state, $\hat{z}(t)=\vec{w}^T \vec{x}(t)$. The weights $\vec{w}$ are adjusted to minimize the time-averaged mean squared error (MSE) between this predicted output and the target output $\vec{z}(t)$.}
\label{fig:basics}
\end{figure}

\section*{Background}
\subsection*{Reservoir Computing}

\label{sec:RC}

In what follows, we present a brief review of reservoir computing in continuous time.  A schematic illustration is provided in \cref{fig:basics}. Readers familiar with RC and measures of capacity may skip to the section labelled Circuit Elements and Structure.

A \emph{reservoir} is a multivariate driven dynamical system. At time $t$, the state of the dynamical system, which we indicate as $\vec{x}(t)$, is driven by an input $\vec u(t)$ and obeys the differential equation
\begin{equation}
    \dot{\vec{x}}(t) = F(\vec{x}(t), \vec u(t)).
\end{equation}
As a result of these dynamics, the state of the reservoir encodes information about $\vec{u}$ as transformations of its previous history. As an instructive example, a linear reservoir is governed by the equation
\begin{equation}\label{eq:linear_reservoir}
    \dot{\vec{x}}(t) = A \vec{x}(t) + \vec u(t).
\end{equation}
which has the long time limit solution
\begin{equation}\label{eq:linear_solution}
\vec x(t) = \int_{0}^\infty d\tau e^{A\tau}\vec{u}(t-\tau)
\end{equation}
showing that the state is a linear function of the past history of $\vec{u}$.  
We will consider reservoirs driven by a scalar input signal $u(t)$ as $\vec u(t)=\vec v u(t)$, where $\vec v$ is a vector of weights that defines how the input signal $u(t)$ enters into each element of the reservoir.

In order for a dynamical system to be considered \emph{feasible} as a reservoir, its state must approach a nontrivial function of the input trajectory in the long time limit. At a high level, we can state this requirement in terms of two conditions:

\begin{itemize}
    \item \emph{Fading Memory}: 
    If the system were to be started from two different initial conditions ${\vec x}_0, \ne {\pvec{x}'}_0$ and driven with the same input trajectory $u$, the system's trajectories should eventually converge to the same state, ${\vec x}(t), {\vec x}'(t) \to {\vec x}[u](t)$ as $t\to \infty$. The statement above implies that the system has a finite temporal memory.
    
    \item \emph{State Separation}: Different input sequences should drive the system into different trajectories, i.e., if the same initial condition were to be driven with two different input trajectories $u \ne  u'$, the resulting reservoir trajectories must be sufficiently different. 
\end{itemize}

The first condition requires that the state of the reservoir becomes a function of the input trajectory, while the second requires that the this function carries information about the input trajectory.

Reservoirs that satisfy the fading memory property have a representation as a Wiener/Volterra series \cite{boyd1985fading},
\begin{align}\label{eq:wiener}
    x_i[u](t) = \int_0^\infty &d\tau_1 \,h_{i1}(\tau_1) u(t-\tau_1) \nonumber +\\
     \int_0^\infty &d\tau_1 d\tau_2 \,h_{i2}(\tau_1, \tau_2) u(t-\tau_1)u(t-\tau_2) + \cdots
\end{align}
which decomposes each degree of freedom $x_i$ into a series of linear and  nonlinear components governed by kernel functions $h_{in}(\tau_1, \dots \tau_n)$. This allows us to regard the reservoir as implementing a \emph{filter} of the input trajectory. Reservoir computers can be thought of as approximating filters, in much the same that feed-forward neural networks may be thought of as approximating functions \cite{maass2002real,maass2007computational}. This will be a useful characterization of both the reservoir and its possible outputs.

In addition to the input trajectory $u(t)$, we are also provided with a target output trajectory $z(t)$. The goal of reservoir computing is to learn to approximate the input-output mapping $u \mapsto z$ 
with an estimate $\hat{z}(t)$, which is a linear combination of the reservoir's variables, $\hat{z}(t) = \vec w^T \vec{x} (t)$. 
Here we will always assume that the $z(t)$ is a scalar. We will also assume that the target output is  a function of $u$ of the form in eqn.~\eqref{eq:wiener}, denoted as $z[u]$. See \cref{fig:basics} for details. 
Finally, as is often done, to the reservoir trajectories $\vec{x}$ we append a constant signal, $\pvec{x}'(t) := [\vec{x}(t), \vec{1}(t)]$ to compensate for constant shifts in the output $z[u]$.

In the following, all quantities depend on the particular input signal $u=\{u(t) : 0\le t\le T\}$ used to drive the reservoir and generate $\vec{x}[u](t)$, $\hat{z}[u](t)$, and $z[u](t)$. This dependence on $u$ can be cumbersome to denote and so we have suppressed it in favor of indicating the dependence on the time interval by the subscript $\square_T$. 

The interesting feature of RC is that only the output layer, given by the coefficients $\vec w$, is trained. Training is performed in the following manner: 
First, the reservoir is ``initialized'' by driving with the input signal
on an interval $[-T', 0]$, until the fading memory property ensures that its state is independent of the initial condition at $t=-T'$. Then, reservoir is driven for an additional interval $[0, T]$. The coefficients $\vec w$ are learned via 
linear regression, by minimizing the time-averaged mean squared error (MSE) between a reconstruction $\hat{z}(t) = \pvec{w}^T \pvec{x}'(t)$ and the target output trajectory $z(t)$ over the time interval $[0, T]$,
\begin{equation}
    \MSE_T[z,\hat{z}] = \frac{1}{T} \int_0^T dt\, (z(t) - \hat{z}(t))^2.
    \label{eq:MSE0}
\end{equation}

This optimization problem $\hat {\vec w}=\operatorname{argmin}_{\vec w} \MSE_T[z,\pvec{w}^T\pvec{x}']$ has a closed form solution (note the use of the `hat' to denote the optimum). Defining the time average, $\langle f \rangle_T = \frac{1}{T}\int_0^T dt\, f(t)$, the solution is ${\hat {\vec w}}=\langle\pvec{x}'^T  
\pvec{x}' \rangle_T^{-1} \langle\pvec{x}' z\rangle_T$ which we show in the supplemental material. Regularization via ridge regression is also commonly employed in practice, which modifies the objective to $\MSE_T[z,\pvec{w}^T\pvec{x}'] +  k||\vec w||^2$ \cite{lukovsevivcius2012practical}.

In the space of functions on $[0, T]$ the optimal approximation $\hat{z}(t)=\hat{\pvec w}^{T} \pvec{x}'(t)$ 
can be understood as the projection of $z(t)$ onto the span of the reservoir trajectories $\pvec{x}'(t)$. This is a consequence of the least squares objective  \cite{dambre2012information}. As the estimate $\hat{z}(t)$ is a projection of $z(t)$, we can evaluate the normalized mean-squared error of the reservoir reconstruction of $z$ as
\begin{equation}
    \text{nMSE}_{T}[z, \hat{z}] = \frac{\MSE_T[z,\hat{z}]}{\langle z^2\rangle_T}.
\end{equation}
It is normalized to $0\le \text{nMSE}_T[z] \le 1$ when calculated on the training interval $t\in [0,T]$.  The $\text{nMSE}$ is often a more useful measure than the $\text{MSE}$, since it gives the \emph{relative} error of the approximation to the variation in $z(t)$.

From now on we will only consider the optimal approximation generated by the reservoir. We introduce the notation
\begin{equation}
    \text{nMSE}_{T}[z] \equiv \min_{\vec{w}} \text{nMSE}_{T}[z, \vec{w}^T \pvec{x}].
\end{equation}
$\text{nMSE}_{T}[z]$ can be read as \emph{the normalized mean-squared error for the reservoir's approximation of $z$ on the time interval $[0, T]$}. Some authors~\cite{dambre2012information} prefer to use a reversed scale, defining the \emph{capacity} of the reservoir to approximate $z$ as
\begin{equation}
    C_T[z] = 1 - \text{nMSE}_{T}[z],
\end{equation}
where $\langle f \rangle_T = \frac{1}{T}\int_0^T dt\, f(t)$.  The capacity is bounded $0 \le C_T [z]\le 1$ with $1$ corresponding to a perfect reconstruction.

As mentioned above, the $\text{nMSE}$ and capacity $C_T$ can be given a geometric interpretation on the space of functions on $[0, T]$.  For a function $z(t)$ normalized as $\langle z^2\rangle_T = 1$, the capacity $C_T[z]$ is the squared length of the projection of $z$ onto the span of the reservoir trajectories $\pvec{x}'$.  The $\text{nMSE}[z]$ is the squared length of the component of $z(t)$ perpendicular to the span of the reservoir trajectories and so $C_T[z] + \text{nMSE}[z]=1$ can be seen as an expression of the Pythagorean Theorem.

\subsection*{Total Memory and Extensivity}

In order to assess a reservoir, we require a more complete picture of its properties, that goes beyond its ability to approximate a single function. A main result of Dambre et al. \cite{dambre2012information} is that the capacities of a reservoir to approximate orthogonal functions give independent information about the reservoir (orthogonality is defined as $\lim_{T\to\infty} \langle zz'\rangle_T \to 0$, see supplemental material for further details).

One consequence of this result is that, in the long-time limit, capacities will converge to time-independent values that characterize the reservoir,
\begin{equation}
    \lim_{T\to\infty} C_T[z] = C[z].
\end{equation}
While we cannot achieve this limit in practice, we can estimate it using techniques from finite size scaling. (The capacities we report throughout this paper are obtained from these techniques, with detailed results of the analysis shown in the supplemental material.)

Another consequence of the result by Dambre is that, rather than evaluating the reservoir's capacity to approximate only a single function $z$, we can evaluate its capacity over a family of orthogonal functions $\{z_i\}$. The resulting capacities allow us to characterize the reservoir's ability to approximate arbitrary  functions of the form $\sum_i a_i z_i$.
We relate the capacities of the reservoir on a family of functions, $C_T[z_i]$, to a linear combination of these functions, $C_T[\sum_i a_i z_i]$, in a precise way:
\emph{Consider a set of $n$ orthonormal functions on $t\in [0, T]$, $\langle z_i z_j\rangle_T = \delta_{ij}$, and normalized linear combination $z = \sum_{i=1}^n a_i z_i$,  $\langle z^2\rangle_T = 1$. Then, if $C_T[z_i] \ge 1-\epsilon$ for all $i=1\dots n$ then $C_T[z] \ge 1-n\epsilon$.}
This is proved in the supplemental material, where we also construct the function $z^*$ that has maximal error. Our construction show that when errors are uncorrelated, or when the maximal error function lies withing the basis, the capacity on the linear combination satisfies the tighter bound $C_T[z^*] \ge 1-\epsilon$.

\begin{figure}[b]
\centering
\includegraphics[width=\columnwidth]{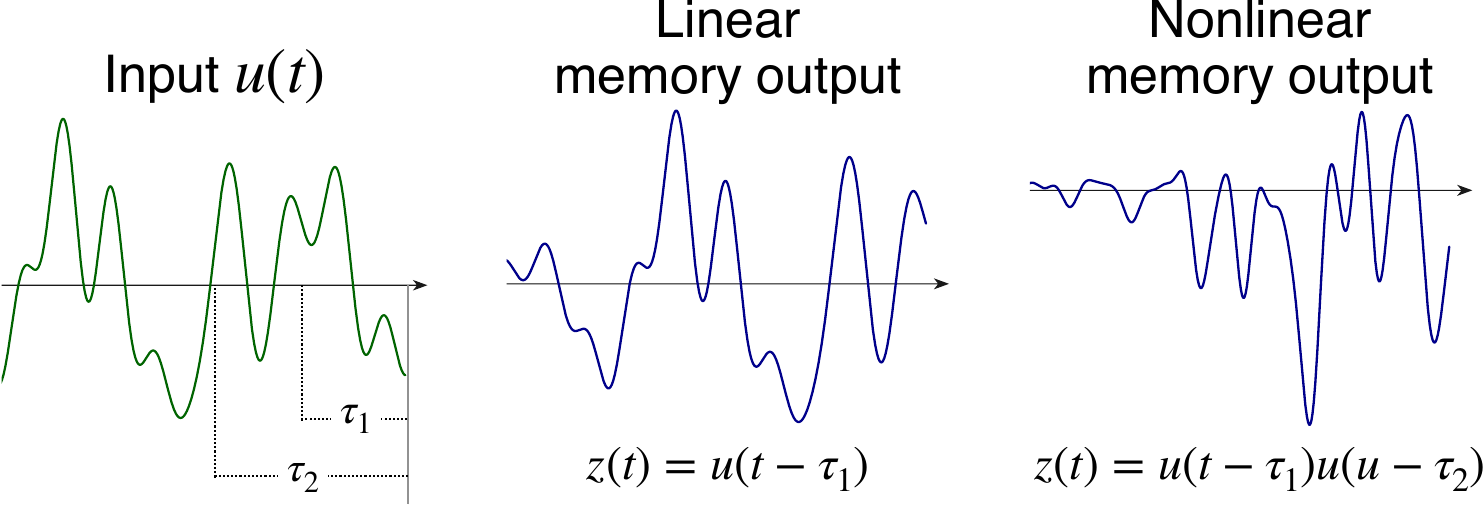}
\caption{The first two terms of the Wiener/Volterra series, eqn.~\eqref{eq:wiener}, are linear and quadratic functions of the input. When taken as the target output $z$, these functions are used to define the linear and nonlinear memory of the reservoirs, as in eqns.~\ref{eq:mlin} and \ref{eq:mnonlin}.}
\label{fig:linnonlin}
\end{figure}

A natural family of functions for evaluating capacities are products of the delayed input $z(t) = u(t-\tau_1)\dots u(t-\tau_n)$, that is  { the terms in the Wiener/Volterra series in} eqn.~\eqref{eq:wiener}.
The first two such functions are shown schematically in \cref{fig:linnonlin}.  
The first  function, $z(t) = u(t-\tau_1)$ leads to the \emph{linear memory function},
\begin{align}
    m(\tau) = C[u(t-\tau)],\label{eq:mlin}
\end{align}
which reflects the ability of the reservoir to reconstruct time-delayed version of the input at different lags~ \cite{hermans2010memory,white2004short,jaeger2002short}. We generalize this concept to nonlinear, $n^{\text{th}}$-order memory functions,
\begin{equation}
  m_n(\tau_1, \tau_2, \dots, \tau_n) = C[u(t-\tau_1)u(t-\tau_2)\dots u(t-\tau_n)],
  \label{eq:mnonlin}
\end{equation}
which reflect the reservoir's ability to approximate the product of the input at various delays in the past. Achieving high nonlinear memory requires the reservoir to  memorize past inputs as well as to be able to perform nonlinear operations on these inputs.

If the input function $u$ is drawn from a family of random functions with mean zero $\langle u\rangle_T = 0$ and decaying auto-correlation $\langle u(t) u(t-\tau)\rangle_T\to 0$ as $\tau \to \infty$, then $u$ and and its time delayed products will approach orthogonality as their delays differ. We generate input functions from smoothed Gaussian noise with correlations that decay exponentially with a unit timescale (see supplemental material). The family of functions $\{u(t-\tau_1),\; u(t-\tau_1)u(t-\tau_2),\; u(t-\tau_1)u(t-\tau_2)u(t-\tau_3)\dots\}$ is approximately orthogonal for time delays satisfying $|\tau_i - \tau_j| > 1$. The memory functions $m_n$ thus tell us independent information about the computational capacities of our reservoir over timescales greater than 1.


Finally, we summarize a reservoir's computational properties via the following quantity:
\begin{equation}
    \tau_\epsilon = \int_0^\infty d\tau\, \Theta (m(\tau) > 1-\epsilon),
\end{equation}
where $\Theta$ is the Heaviside step function. We refer to $\tau_\epsilon$ as the \emph{total linear memory} of the reservoir. This quantity captures the time delay up to  which the history of the input is reconstructed with an error less than $\epsilon$.
This definition can be generalized to quantify the \emph{total $n^{\text{th}}$-order nonlinear memory} as
\begin{equation}
   \tau^{(n)}_\epsilon = \int_0^\infty\cdots \int_0^\infty d\tau_1\cdots d\tau_n\, \Theta (m_n(\tau_1, \dots, \tau_n) > 1-\epsilon).
\end{equation}
Previous studies have demonstrated linear reservoirs with extensive total memory, \emph{i.e.} $\tau_\epsilon \propto N$ for a reservoir with $N$ elements~\cite{white2004short,hermans2010memory}.  In this work we present an electronic implementation of the linear reservoirs discussed in those papers. We then  generalize this approach by designing an electronic reservoir which displays extensive scaling in its total  quadratic-memory, $\tau^{(2)}_\epsilon \propto N$.

\subsection*{Circuit Elements and Structure}

\begin{figure}
\centering
\includegraphics[width=\columnwidth]{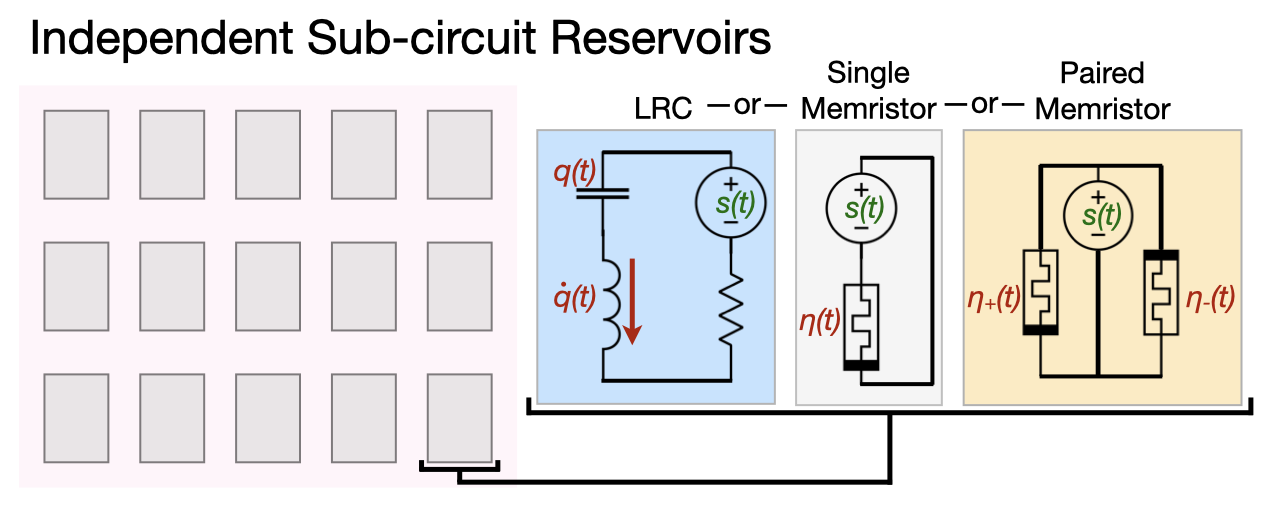}
\caption{A class of reservoirs we consider are composed of independent sub-circuits. Each circuit has identical structure, but with parameters that vary to achieve independent trajectories when driven with the same input. In all sub-circuits the input is a voltage generator driven with $s(t)$ and shown in green. The output circuit variables are shown in red. The three sub-circuit types are: LRC circuits with output variables $q(t)$, the charge on the capacitor and $\dot{q}(t)$, the current through the inductor, single memristors with output variable $\memstate(t)$, the memristive state variable, and paired memristors biased in opposite directions with output variables $\memstate_+(t)$ and $\memstate_-(t)$. As the resistance is a linear function of $\memstate$, the two are equivalent as output variables.}
\label{fig:subcircuitreservoirs}
\end{figure}

\begin{figure}
\centering
\includegraphics[width=0.8\columnwidth]{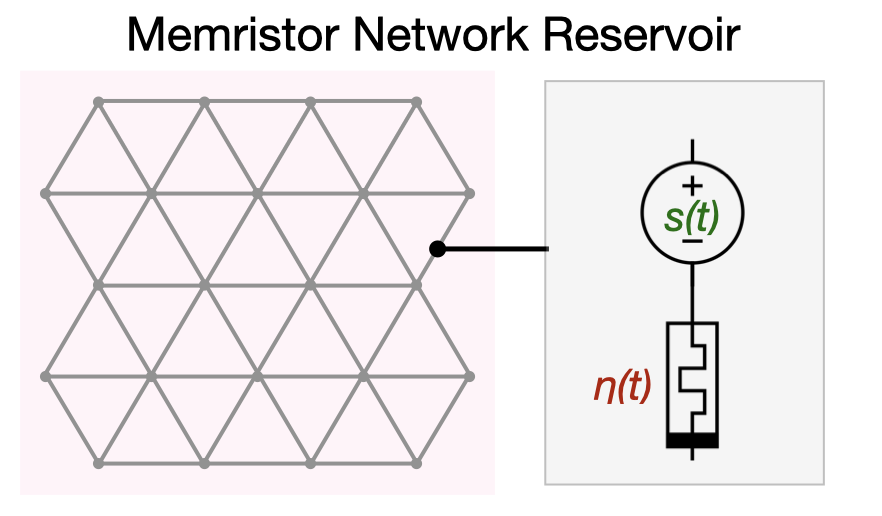}
\caption{The edges of the memristor networks we consider consist of voltage driven memristors which are arranged in a triangular lattice. The output variables are the memristive states in the networks $\vec{\memstate}(t) = [\memstate_1(t), \memstate_2(t), \dots ]$. The magnitude of the driving signal in each edge $s_e(t)$ varies on the interval $[S, -S]$ in equally spaced increments. }
\label{fig:memnet}
\end{figure}

We consider electronic circuits composed of traditional linear elements including inductors (L), capacitors (C), and resistors (R), active elements (voltage or current sources), as well as passive memory elements known as memristors (Mem) (see below).  In all cases, the electronic reservoir will accept a vector input through a set of voltage sources $\vec{s}$.  We can convert the scalar input $u(t)$ to a time dependent voltage vector $\vec{s}(t)$ through a set of constant input weights $\vec{s}(t) = \vec{v}u(t)$. We will consider several circuit structures as shown in \cref{fig:subcircuitreservoirs,fig:memnet}, including independent sub-circuit reservoirs composed of LRC, single and paired memristor circuits and memristive networks on a lattice.  

The linear reservoirs we consider will be composed of sets of LRC sub-circuits as shown in \cref{fig:subcircuitreservoirs} in which each LRC circuit, indexed by $n$ has component values $l_n,\, r_n, c_n$ and is driven by a voltage generator $s_n = u(t)$ (taking $\vec{v} = \vec{1}$).  Each circuit possesses two degrees of freedom $q_n(t), \, \dot{q}_n(t)$  corresponding to the charge across and current entering the capacitor, which obey the following equations of motion:
\begin{equation}\label{eq:lrcqqdot}
    \begin{bmatrix} \dot{q}_n(t) \\
    \ddot{q}_n(t)\end{bmatrix} = 
    \begin{bmatrix}
    0 & 1 \\
    -\frac{1}{l_n c_n} & -\frac{r_n}{l_n}
    \end{bmatrix}
    \begin{bmatrix}
    q_n(t) \\
    \dot{q}_n (t)
    \end{bmatrix}
    + \begin{bmatrix}
    0 \\
    \frac{s_n(t)}{l_n}.
    \end{bmatrix}
\end{equation}
The output trajectories of an LRC sub-circuit are the trajectories of the internal degrees of freedom, $\vec{x} = [\vec{q}(t), \dot{\vec{q}}(t)]$ (the vector notation covers the indexing over $n$).  The dimension of an LRC reservoir of $N$ sub-circuits is thus $2N$. In the supplemental information we prove that a network of LRC motifs is equivalent to a collection of separate sub-circuits. There is thus no benefit to considering a network of these motifs and our treatment here is fully general.

In addition to linear elements, we consider reservoirs of nonlinear electronic components called memristors.
Memristors are passive 2-terminal devices characterized by the response relationship,
\begin{align}
    V(t) &= R(\memstate)I(t), \label{eq:z1}\\
    \dot{\memstate}(t) &= f(\memstate(t), I(t)),\label{eq:z2}
\end{align}
where $V(t)$ is the voltage drop across the memristor,
$I(t)$ is current, $\memstate(t)$ is the internal state of the memristor, and  $R(\memstate(t))$ is the state-dependent resistance. It can be seen that the resistance can depend on the past history of the current.  Importantly, memristors are inherently nonlinear elements. As in the case of linear networks, the input to the circuits is through a set of voltage generators $\vec{s}(t) = \vec{v}u(t)$ where $\vec{v}$ is a constant vector of weights with units of voltage.

The internal states of a memristor circuit closely mimic the behavior of a neural network. We constrain ourselves to the linear current model similar to the one proposed in \cite{strukov2008missing}, along with a decay term \cite{caravelli2017complex},
\begin{align}
    R(\memstate) &= \Roff (1-\memstate) + \Ron \memstate,\label{eq:a000}\\
    R(\memstate) &= \Roff (1-\chi \memstate)\quad\quad\qquad \Big( \chi := \frac{\Roff  - \Ron }{\Roff } \Big), \\
    \dot{\memstate}(t) &= -\alpha \memstate(t) + \frac{\Roff}{\beta} I(t).\label{eq:memristor_eom}
\end{align}
Here the constant $\alpha=1/{t^*}$ is an inverse time scale while $\beta$ is an activation current per unit of time, which moderates the strength of the input signal. This model is the simplest approximation of a current-controlled memristor, as we show in the supplementary material. The internal state $\memstate$ is limited to the interval $[0, 1]$ by hard barriers, so the resistance  $R(\memstate)$ varies between two limiting values $[\Ron ,\Roff ]$. We use $\memstate$ as the output variable of memristors, but as the resistance $R$ is a linear function of $\memstate$, the two are equivalent as components of a linear regression.

The term $-\alpha\memstate(t)$ in eqn.~\eqref{eq:memristor_eom} causes the memristor state $\memstate$ to decay to 0 in the absence of a current.  This is called volatility in the context of memristors and is an important effect in many memory materials and their applications (see the review \cite{wang2020volatile}). Here it plays a central role in providing the fading memory property for memristor circuits. However, not all memristors are volatile and commercial memristors are generally designed to be nonvolatile for memory applications. While nonvolatile memristors can still show fading memory (see \cite{ascoli2016history,menzel2017origin}), nonvolatile fading memory is much more difficult to treat analytically and so we limit our study to the case of volatile memristors. This is not as great a limitation as it might seem, in the sense that any nonvolatile memristor can be turned into a volatile memristor by the addition of a voltage or current source that gives a small negative current in the absence of an applied voltage.

When considering networks of elements as in \cref{fig:memnet} it is convenient to introduce the cycle space projector $\Omega_A$ \cite{caravelli2017complex}, which projects a vector that a assigns a real number to each edge of a graph onto current configurations that satisfy Kirchhoff's current law. The projector $\Omega_A$ can be simply computed from the circuit graph $\mathcal G$, in which nodes of the graph represent  electrical junctures and edges contain electrical elements.
For a circuit in which the edges contain a voltage generator in series with a memristor with values $\vec{u}(t)$, as in \cref{fig:memnet}, 
the equation of motion is given by \cite{caravelli2017complex}
\begin{equation}\label{eq:memnet}
    \dot{\vec{\memstate}}(t) = -\alpha \vec{\memstate}(t) + \frac{1}{\beta} (I - \chi \Omega_A H(t))^{-1} \Omega_A \vec u(t)
\end{equation}
where we have used the convention $\Memstate(t) = \mathrm{diag}\, \vec{\memstate}(t)$.  Interactions between memristors thus occur through the inverse of $I -\chi\Omega_A \Memstate(t)$ and are mediated both by $\chi$ and the Kirchhoff laws imposed by $\Omega_A$.
A memristor circuit generates reservoir trajectories $\vec{x} = \vec{\memstate}(t)$.

\section*{Results}

\subsection*{Linear Electronic Reservoirs}

\begin{figure}
\begin{center}
    \includegraphics[width=8.6cm]{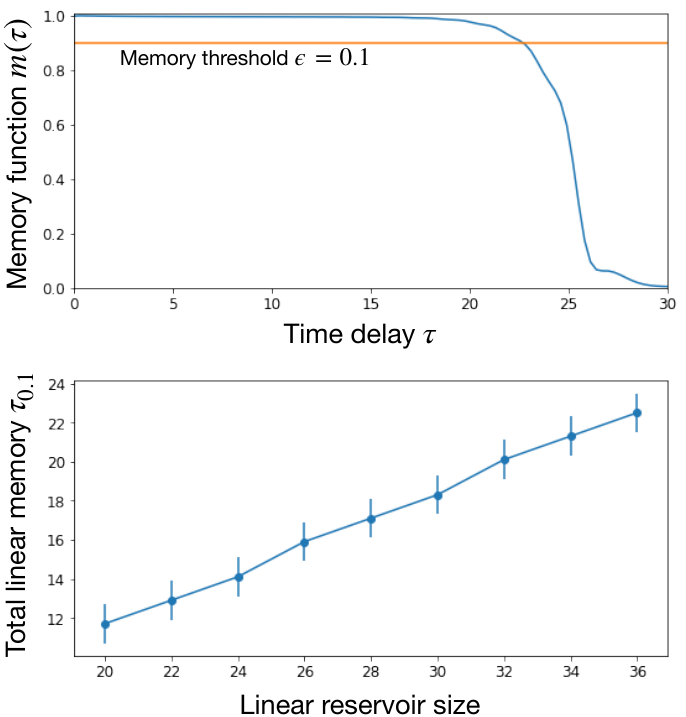}
\end{center}
\caption{The top panel displays the memory function for an LRC reservoir of 18 sub-circuits (36 trajectories) with $\Delta\omega = \gamma = \frac{4}{18}$.  The memory function $m(\tau)$ displays the capacity of the reservoir to reconstruct the delayed input $u(t-\tau)$.  The reservoir retains an accurate memory of the input before falling sharply. In the bottom panel we display the $\epsilon=0.1$ total linear memory of LRC networks of various sizes.  This corresponds to the signal delay at which the memory function falls below 0.9, displayed in the top panel as a vertical line.  As discussed in the main text, the total linear memory scales extensively with the system size. 
 \label{fig:memfunc_scaling}}
\end{figure}

In this section we illustrate the ideas of memory and scaling introduced above in the familiar realm of linear circuits. We show that electronic reservoirs with extensive total memory can be constructed from LRC circuits and that these circuits can be understood as calculating a time-windowed version of the Fourier transform, akin to a spectrogram. Moreover, in the supplemental material, we prove the following result:\ \\

\noindent \textbf{Feasibility of LRC circuits}. \textit{Reservoirs of LRC circuit motifs satisfy the fading memory and state separation properties.}\ \\

This justifies our use  of these systems as reservoirs and the application of the memory measures defined above.

From the solution in eqn.~\eqref{eq:linear_solution}, we observe that linear reservoirs generate linear functions of the input signal, and that their properties will depend on the eigenvalues of $A$.  We thus examine the scaling of the total linear memory $\tau_\epsilon$ using spectral techniques. In \cite{white2004short,hermans2010memory}, it was noted that an extensive total memory was obtained for reservoirs with eigenvalues lying on a vertical line in the negative half plane, {i.e.,} eigenvalues of the form $\lambda_{n, \pm} = -\gamma \pm i n \Delta \omega$ for $n\in \{0 \dots N\}$.  
As shown in the supplemental material, the solution to eqn.~\eqref{eq:lrcqqdot} depends on linear combinations of the integrals $\int_0^\infty d\tau \,e^{\lambda_{n,\pm}\tau} u(t-\tau)$. For the values of $\lambda_{n,\pm}$ above, this  can be interpreted as calculating a local Fourier transform of the input. The frequencies $\textrm{Im}(\lambda_{n,\pm}) = \pm n\Delta \omega$ are evenly spaced with largest frequency $ N\Delta \omega$ and smallest frequency $\Delta \omega$.  The exponential window, $e^{-\gamma\tau}$ applies a cutoff in time on the interval $0\le \tau\le 1/\gamma$. This means that in order to not experience interference with previous time windows, we should set the lowest frequency, $\Delta\omega$ to be on the same order as $\gamma$. Further details on this correspondence are given in the supplemental material (section VIII. Solution of LRC Circuits).

The LRC sub-circuit in \cref{fig:subcircuitreservoirs} and governed by eqn.~\eqref{eq:lrcqqdot} has eigenvalues $\lambda_\pm = -\frac{r}{2l} \pm i \sqrt{\frac{1}{lc} - \frac{r^2}{4l^2}} = -\gamma \pm i \omega$ and a corresponding pair of trajectories $q_n(t),\, \dot{q}_n(t)$ corresponding to the charge and current entering the capacitor  (see supplemental material).   As a consequence \emph{any eigenvalue spectrum in the negative half-plane and symmetric in the upper and lower half planes can be achieved by a collection of LRC circuits with a particular choice of the component values $l_n, r_n, c_n$}.
Given a $\gamma$ and $\Delta\omega$, we choose $l_n = 1$, $r_n = 2\gamma$ for all $n$ and
\begin{equation} \label{eq:lrccomp}
    c_n = \frac{1}{n^2 \Delta\omega^2 + \gamma^2},
\end{equation}
in which case the resulting LRC circuits have eigenvalues $\lambda_{n,\pm} = -\gamma \pm i n\Delta\omega$. 

Reconstructing the input at a given time lag, $z(t)=u(t-\tau_1)$, is equivalent to constructing an approximate representation of the delta function as the linear term of eqn.~\eqref{eq:wiener},  $h_1(\tau) = \delta(\tau_1-\tau)$.  After fitting, the learned weights $w_{q_n}$, $w_{\dot{q}_n}$ and $w_c$ may be used to construct the kernel $h_1(\tau)$ of the reservoir. An example of this is given in the supplemental material, which makes it clear that the training procedure is in fact constructing a delta function approximation.

To construct an LRC network of $N$ circuits, we identify a maximum frequency $\omega_{\max}$ associated with our input signal and define $\Delta\omega = {\omega_{\max}}/{N}$ as the lowest frequency and resolution.  This lowest frequency defines a timescale $t'\sim {1}/{\Delta \omega}$ over which a signal could be represented by the Fourier series.  We then choose $\gamma = \Delta\omega$ to suppress the signal for times longer than $t'$. For the smoothed Gaussian noise input signal used here (see supplemental material), we take $\omega_{max} = 4$.

We expect that if $\omega_{\max}$ is chosen to be sufficiently large so as to accurately represent the signal, the system's total linear memory $\tau_{\epsilon}$ will scale as $N$.  This is confirmed in \cref{fig:memfunc_scaling}, where we show the memory function $m(\tau) = C[u(t-\tau)]$ and the total linear memory $\tau_\epsilon$ for $\epsilon=0.1$ across a range of reservoir sizes.  As expected, reservoirs of this type show an extensive total linear memory.  Note that the precise value of $\epsilon$ is not particularly important since the memory function for these networks maintains a value near 1 before falling sharply. This implies that  an LRC network will be able to approximate any function $z(t) = \int_0^{T^*} d\tau\, h_1(\tau) u(t-\tau)$ for any kernel $h_1$, so long as the network is large enough that its total linear memory obeys $\tau_\epsilon > T^*$.

\section*{Memristor Reservoirs}

We now turn to establishing analogous results for memristor reservoirs.  To begin, in the supplemental material we prove the following result:\ \\

\noindent \textbf{Feasibility of Memristor Networks}. \textit{Reservoirs of networked memristors satisfy local fading memory and state separation properties for sufficiently weak input signal.}\ \\

Specifically, the proof requires that the driving signal satisfy $||\vec{u}(t)||_2 \le \frac{(1-\chi)^2 \alpha \beta}{\chi}$. This result was a surprise given the passive nature of the memristor model but simulations have demonstrated the importance of weak driving to the fading memory property. The dependence on $\chi$ illustrates the constraints put on the driving signal by the nonlinearity of the memristors. The condition above illustrates a tradeoff between nonlinearity and volatility, governed by the decay constant $\alpha$ in maintaining the fading memory property.  We also note that nonvolatile memristors have demonstrated a form of fading memory \cite{ascoli2016history,menzel2017origin}, though it is not clear what is at the root of this effect.  Certainly, saturation effects at the boundary of the devices can also lead to fading memory for a strong signal and this will be in the opposite regime to the bound above. It is certainly possible that different memristors show fading memory over a wider range of driving signals and for varying reasons, but our investigations have made it clear that the linear model will only satisfy the local fading memory property when weakly driven. In the supplemental material we also show that, so long at the dissipative term is bounded below by a linear function of $\vec\memstate$, a very similar bound will hold.

{Further work has shown that when strongly driven, memristor networks can enter a transiently chaotic state in which trajectories will diverge in time \cite{caravelli2021global}, in apparent violation of the local fading memory property. However, after this transiently chaotic phase of the dynamics, the system will approach a steady state under constant driving. Presently it is not clear whether systems that display transient chaos can form useful reservoirs when the driving signal is time-varying, as chaotic dynamics may recur repeatedly. Our simulations have indicated that maintaining the fading memory property is important, so all memristor networks we employ are in the weakly driven regime.}

The trajectory of a single memristor governed by eqn.~(\ref{eq:memristor_eom}) has the Volterra series expansion
\begin{multline}
\memstate(t) =  \frac{1}{\beta}\int_0^\infty \!\! d\tau_1 e^{-\alpha\tau_1 } u(t-\tau_1) + \\ \frac{\chi}{\beta^2} \int_0^\infty \!\! d\tau_1 \int_{\tau_1}^\infty\!\!  d\tau_2\, e^{-\alpha\tau_2} u(t-\tau_1) u(t-\tau_2) +O(\chi^2),
\label{eq:volt0}
\end{multline}
(see supplemental material), where we have neglected boundary effects and where higher order kernels of the input are suppressed by higher powers of $\chi$.  We immediately note that the kernel functions are exponentially decaying at a rate $\alpha$. This is consistent with the fading memory property, as any perturbation in $u$ will have an exponentially decreasing effect on $\memstate$ as time goes on. The expansion makes it clear that memristor trajectories depend on higher powers of the driving signal $u(t)$.  In this sense, memristors may be considered as a source of nonlinearity in electronic reservoirs.

As a generalization of the linear approximation task, we consider the approximation of an arbitrary 2nd order filter which depends on the input signal for a time $T^*$ into the past. Specifically, we wish to approximate a function of the input with the form,
\begin{multline} \label{eq:2ndorder}
    z(t) = \int_0^{T^*}\!\! d\tau_1\, F_1(\tau_1) u(t-\tau_1) + \\
    \int_0^{T^*} \!\! d\tau_1 \int_0^{T^*} \!\! d\tau_2\, F_2(\tau_1, \tau_2) u(t-\tau_1)u(t-\tau_2).
\end{multline}
Approximating this function with memristors requires isolating terms related to the second order contribution to the trajectory in eqn.~\eqref{eq:volt0}.  To do this, we note that a reservoir of two memristors $\memstate_+(t)$ and $\memstate_-(t)$, each driven by $u(t)$ and $-u(t)$ respectively, allows us to construct their sum (away from the boundaries and up to terms of order $O(\chi^2)$) as
\begin{align*}
    \memstate_+(t) + \memstate_-(t) &\approx \\
    \frac{2\chi}{\beta^2} \int_0^\infty & d\tau_1  \int_{\tau_1}^\infty  d\tau_2\, e^{-\alpha\tau_2} u(t-\tau_1) u(t-\tau_2), 
\end{align*}
which cancels all odd terms in $u(t)$ in eqn. \eqref{eq:volt0}. Including memristors in pairs thus allows the training procedure to isolate these quadratic components and learn their weights so as to approximate eqn.~\eqref{eq:2ndorder}.

\begin{figure*}
\begin{center}
    \includegraphics[width=17.8cm]{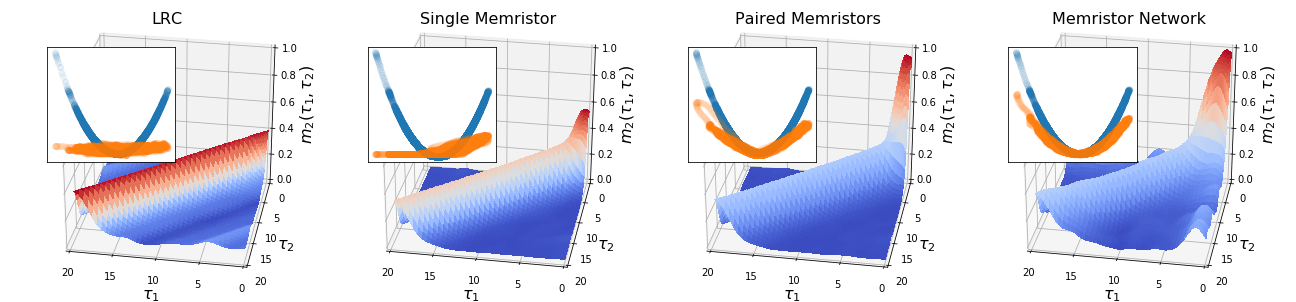}
\end{center}
\caption{  
The quadratic memory function $m_2(\tau_1, \tau_2)$ for a single LRC sub-circuit (left), a single memristor sub-circuit (middle left), apaired memristors sub-circuit (middle right) and a triangular lattice of memristors (right) as shown in \cref{fig:subcircuitreservoirs,fig:memnet}.  This measures the reservoirs ability to approximate the function $z(t) = u(t-\tau_1)u(t-\tau_2)$. 
The colors of the surfaces are purely for visualization and values should be read from the z-axes which all extend from 0 to 1. We define $\tau^*:= \text{argmax}_\tau m_2(\tau,\tau)$, the optimal delay for an equal-time reconstruction; the insets show the corresponding reconstruction $\hat{z}$ (orange) and target output $z(t)=[u(t-\tau^*)]^2$ (blue) as a function of the input signal $u(t-\tau^*)$. As expected, the LRC circuit is only capable of generating linear approximations of the output. A single memristor reservoir is unable to isolate its quadratic component and misses negative parts of the reconstruction due to boundary effects (middle left inset).  The addition of another memristor with opposite bias, significantly increases the ability to reconstruct $z$.  The memristor network shows an enhanced ability to reconstruct $z$ and clear nonlinearity (right inset). The nonlinear memory function value of $m_2(\tau^*, \tau^*)$ for each network was 0.390 (LRC), 0.558 (single memristor), 0.960 (paired memristors), and 0.995 (memristor network).}
\label{fig:memquadcap}
\end{figure*}

The $e^{-\alpha\tau}$ dependence of terms in the Wiener/Volterra series  only allows a dependence on $u$ on timescales of order $\frac{1}{\alpha}$. A lower value of $\alpha$ will integrate a longer window of the previous history into the current state, but will also obscure the value of the input signal at any single time. In reservoir computing, the parameters of the circuit are randomized to generate linearly independent trajectories, which allows the training to isolate different components of the input signal.
In memristor networks, this may be accomplished by varying $\alpha$ and $\beta$ (the timescales of decay/excitation for memristors), by varying the amplitude of the driving, or by introducing disorder into the structure of the circuit $\Omega_A$. We employ  what we view as the most practical option, which is varying the amplitude of the driving signal.  In networks, memristors are driven with a proximal voltage generator that varies in amplitude from $+S$ to $-S$ in equally spaced increments,  where $S$ is a constant that may be tuned.

In \cref{fig:memquadcap} we show the quadratic memory function $m_2(\tau_1, \tau_2) = C[u(t-\tau_1)u(t-\tau_2)]$ for reservoirs composed of LRC, single memristors, paired memristors and memristor networks as shown in \cref{fig:subcircuitreservoirs,fig:memnet}. As expected, while the LRC reservoir produces excellent linear reconstructions, it shows very poor ability to reconstruct quadratic functions. Among memristors while single memristors show clear nonlinearity, paired memrisitors are markedly better and memristor networks show only a limited advantage over separate paired memristor sub-circuits.
  
While memristor reservoirs give us the ability to calculate quadratic functions of the input with high accuracy for short times, the total quadratic memory  $\tau^{(2)}_\epsilon$ does not scale extensively as the size of the reservoir is increased. This can be seen from the fact that the memristor network reservoir, which is 28 times bigger than the ``paired memristors'' reservoir, has a similar total quadratic memory. In the next section we consider hybrid reservoirs of memristor and LRC components, which do demonstrate extensive scaling.

\subsection*{Hybrid Deep Reservoirs}

\begin{figure}[b]
    \centering
    \includegraphics[width=8.6cm]{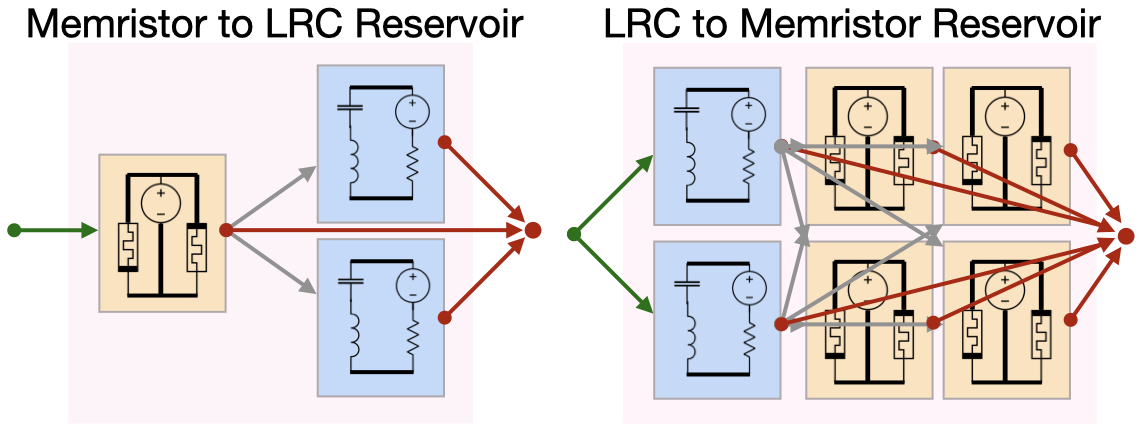}
    \caption{We consider two deep reservoir structures in which the layers are independent sub-circuit reservoirs of LRC or memristor sub-circuits. Connections and layers are shown schematically to make the figures legible. Inputs to the reservoirs are shown in green, internal connections between layers are shown in grey and outputs are shown in red. In the memristor to LRC reservoirs (left), the surface layer is a paired memristor circuit. Each of the output trajectories $\memstate_\pm(t)$ is used to drive a deep layer reservoir of 10 LRC sub-circuits. In the LRC to memristor reservoirs (right) each pair of surface layer LRC circuits drives 12 paired memristor circuits (not all shown here) such that all sums and differences of the 4 LRC trajectories drive a separate paired memristor circuit.
    }
    \label{fig:hybridcircuit}
\end{figure}

\begin{figure}
\begin{center}
    \includegraphics[width=8.6cm]{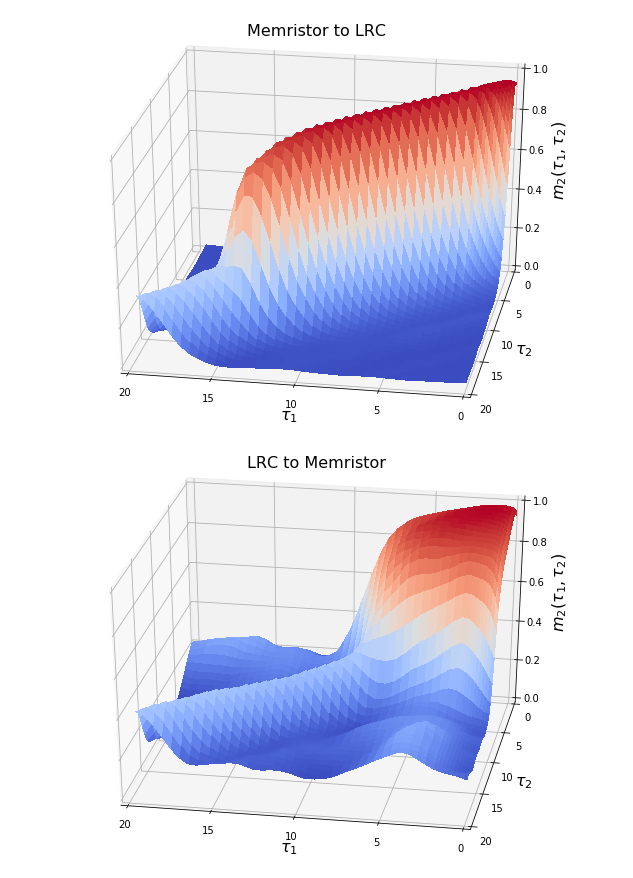}
\end{center}
\caption{
The quadratic memory function $m_2(\tau_1, \tau_2)$ for the hybrid memristor and LRC reservoirs shown in \cref{fig:hybridcircuit}.  The top panel shows the result of using a driven pair of memristors to drive an LRC reservoir.  The LRC reservoir stores memory of the nonlinear computation in the memristor network, leading to large equal-time quadratic capacities.  In the lower panel, the result of using an LRC reservoir to drive a set of memristor pairs is shown.  The memristor pairs compute products of the trajectories generated in the LRC network, approximately implementing a 2-dimensional Fourier transform. This extends the quadratic memory function to longer delays compared with the paired memristor reservoir in \cref{fig:memquadcap}. The LRC circuits were arranged with $\gamma = 0.4$, $\Delta\omega = 0.4$ corresponding to a cutoff frequency of 4 $Hz$ \label{fig:hybrid}}
\end{figure}

\begin{figure}
    \centering
    \includegraphics[width=8.6cm]{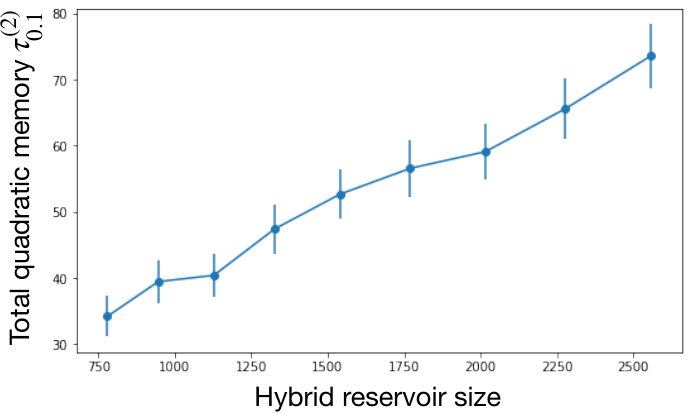}
    \caption{The scaling of the total quadratic memory  with reservoir size in the hybrid LRC to Memristor reservoir.  LRC reservoirs ranging from 10 to 18 sub-circuits were used to drive a memristor reservoir as described in the main text, resulting in reservoirs ranging from 946 to 2556 internal degrees of freedom.  The total quadratic memory  $\tau^{(2)}_{0.1}$ (and errorbars) were estimated from a finite size scaling analysis as detailed in the supplemental material.
    }
    \label{fig:scaling}
\end{figure}

The properties of LRC and memristor reservoirs may be combined to achieve improved scaling of the nonlinear memory. The reservoir structure we will examine uses the trajectories of a ``surface layer'' reservoir $\vec{x}_s(t)$ where voltage generators are driven by the input, $\vec{s}_s = \vec{v}u(t)$, to drive another ``deep layer'' reservoir  $\vec{x}_d$ \cite{deeprc}. The deep layer voltage generators are driven by the surface layer trajectories as $\vec{s}_d = C\vec{x}_s$ where $C$ is a matrix of coupling coefficients whose structure is discussed below. As LRC and memristor components are kept in separate layers, these deep reservoirs inherit the feasibility properties of their sub-components.  The training procedure uses all trajectories $\vec{x} = [\vec{x}_s, \vec{x}_d ]$ in the regression.

As seen in \cref{fig:memquadcap}, the ability of memristors to calculate quadratic functions of the input occurs only over very short delays. On the other hand, LRC networks show excellent linear memory approximations but cannot reconstruct nonlinear functions.
This would suggest that using a surface layer memristor network to generate nonlinear transformations of the input, and then using these to drive a deep layer LRC reservoir that would remember them, would give both good quadratic reconstructions and long memory of these reconstructions. In \cref{fig:hybrid}, the top panel shows the result of using a pair of memristors configured as in the section above, to drive an LRC reservoir. Each of the two memristor trajectories in $\vec{x}_s = [\memstate_+(t),\, \memstate_-(t)]$ is used as a source signal for a small LRC reservoir of 10 circuits,  which we index by $n\pm$ with $n=1\dots 10$. Each LRC circuit produces 2 trajectories $q_{n\pm}, \dot{q}_{n\pm}$ giving a total of $2+2\times10\times2=42$ output trajectories.  The LRC trajectories are calculated by eqn. \eqref{eq:lrcqqdot} with $s_{n\pm} = \memstate_\pm$. The index $n$ determines the parameters of the LRC elements given $\gamma=0.4$ and $\Delta\omega=0.4$ and eqn. \eqref{eq:lrccomp}.

The resulting reservoir trajectories are used to evaluate the quadratic memory function in the top panel of \cref{fig:hybrid} with the results showing a substantial increase in the reservoir's computational capacity for `equal-time' quadratic reconstructions (defined via $m_2(\tau,\tau)$).  As the deep LRC reservoir is used to recall the equal-time products computed by the memristor reservoir, we expect that measures of their total quadratic memory $\tau^{(2)}_\epsilon$ will also scale extensively.  However increasing the reservoir size will not  improve the reconstruction of unequal-time products where $\tau_1\neq \tau_2$.

We next consider using a surface layer LRC reservoir to drive a deep layer memristor reservoir. As guiding intuition, if we consider the LRC reservoir as computing the Fourier transform of the signal, the deep memristor layer will calculate products of this transform, akin to a 2-dimensional Fourier transform in $\tau_1$ and $\tau_2$.  We predict that the resulting network will display an improved unequal-time quadratic memory function.  To test this, we implemented the same 10 circuit LRC reservoir as described above driven with the same input signal $u$.  The resulting 20 trajectories, $q_n, \dot{q}_n$, $n=1\dots 10$ are used to drive a set of memristor pairs such that the sum and difference of every pair of the 20 LRC trajectories are used to drive an independent pair of memristors. This means that for a particular pair $\memstate_{m+}, \memstate_{m-}$, the driving signal may be $q_n \pm q_{n'}$, $q_n \pm \dot{q}_{n'}$ or $\dot{q}_n \pm \dot{q}_{n'}$ such that all pairs $n,n'$ and $q,\dot{q}$ are used.  The resulting $2\times 20\times 19+20 = 780$ trajectories are used to train the reservoir.  In the lower panel of \cref{fig:hybrid}, we calculate the quadratic memory function for this architecture.  We observe a substantial improvement in the reservoir's ability to construct unequal-time products of the input signal. Although this requires a significant increase in the size of the reservoir, such an increase is expected. The number of unequal-time products with $\tau_1, \tau_2 < T^*$ scales quadratically in the maximum delay $T^*$ and so the reservoir size must scale similarly.

In \cref{fig:scaling} we show the scaling of the total quadratic memory $\tau^{(2)}_{0.1}$ with the size of the reservoir.  The total quadratic memory indeed scales extensively with the size of the reservoir indicating that arbitrary unequal-time products may be reconstructed by a sufficiently large reservoir. Estimating these quantities accurately turns out to be quite subtle, as estimated values will display strong bias when calculated on a finite interval. In the supplemental material we show how finite size scaling can be used to obtain reliable estimates. We emphasize that it is the total nonlinear memory $\tau^{(2)}_\epsilon$ that can scale extensively with the system size; increasing the maximum delay $T^*$ under which we can reconstruct products of the input will require that the reservoir size scale as ${T^*}^2$.

Another natural architecture to consider would use memristor reservoirs as both surface and deep layers, as has been considered in works based on the simulation of these devices.  Given the discussion above, we expect the primary benefit of this architecture would be to enhance higher order nonlinear capacities, which is precisely what we observe in simulation.  However, as this is outside the scope of the computational task we set out to achieve, we do not include results from such networks here.

\subsection*{Comparison with Echo State Networks}

To show that these design considerations lead to improved performance, we construct a fitting task in which we must approximate a known function of the input signal.  We construct the following target output:
\begin{align}\label{eq:comptask}
    z(t) &= \int_0^{10} d\tau\, K_1(\tau_1) u(t-\tau_1) + \nonumber\\
    {}& \int_0^{10} d\tau_1 \int_0^{10} d\tau_2 K_2(\tau_1, \tau_2)u(t-\tau_1) u(t-\tau_2)
\end{align}
where the kernels $K_1$ and $K_2$ are defined as 
\begin{align}
    K_1(\tau_1) &= e^{-0.5 \tau_1}\cos(2\tau_1) \\
    K_2(\tau_1,\tau_2) &= -e^{-0.3(\tau_1+\tau_2)}\cos\big(2(\tau_1 - \tau_2)\big).
\end{align}
Accurate approximation of this output requires a mixture of memory and nonlinearity.


In addition to the hybrid reservoir discussed above, we also apply an implementation of continuous time Echo State Networks (ESNs)\cite{jaeger2001echo,lukovsevivcius2012practical} for comparison. An ESN is a dynamical reservoir in which the internal states evolve according to
\begin{eqnarray}
    \dot{\vec x} =-\alpha {\vec x}(t)+\tanh(M {\vec x}(t) + \vec v u(t)),
\end{eqnarray}
where $\alpha$ is a decay term,  $\tanh(\cdot)$ applies to every neuron, $M$ is a matrix that in order to satisfy the fading property must have maximum eigenvalue less than one, and $\vec v$ scales the magnitude of the input $u(t)$ which drives each neuron.

We compare the results of a suitably tuned ESN
to a pure memristor network as in \cref{fig:memnet}, and a hybrid surface LRC to deep memristor reservoir. The memristor network  is a $17\times 17$ triangular lattice with 800 edges each containing a memristor ($\alpha = 3$, $\beta = 1$, $\chi = 0.8$) and voltage generator.  The elements of $\vec{v}$ were uniformly distributed on the interval $[-1, -0.1]\cup [0.1, 1]$. The  LRC$\to$Memristor reservoir was configured identically to that presented in the previous section and included a total of 780 trajectories. Finally the ESN consisting of 780 elements was run and tuned following the recommendations in \cite{lukovsevivcius2012practical}. 
As far as possible, each reservoir was configured to produce the same number of trajectories (the lattice structure of the naive memristor network imposes some constraints). Further details on the implementation of each reservoir can be found in the supplemental material. The  LRC$\to$Memristor reservoir was configured identically to that presented in the previous section and included a total of 780 trajectories. Finally the ESN consisting of 780 elements was run and tuned following the recommendations in \cite{lukovsevivcius2012practical}.

Each reservoir was first initialized on the interval $[0, 100]$ corresponding to approximately 100 autocorrelation times of the input driving signal. They were then trained on the interval $[100, 4000]$ and the $\textrm{nMSE}_{[100,4000]}$ is reported in table \ref{tab:comp}. Lastly, a generalization error is reported by calculating the $\textrm{nMSE}[z]$ on the interval $[4000, 5000]$ with the weights that were trained on $[100, 4000]$. The generalization error is no longer normalized to $[0, 1]$ as the weights are calculated on a different interval. However it is the most important measure of the reservoir's ability to approximate the function $u \mapsto z$, rather than to simply fit this function on a single training interval.

The results of the training are shown in table \ref{tab:comp}. The hybrid LRC$\to$Memristor reservoir demonstrates a 10-fold improvement over the pure memristor network and performs on par with the ESN implementation in training as well as a 2-fold improvement in generalization error.  We attribute this to the specialized structure of the LRC to memristor reservoir, which gives it an advantage at reliably calculating quadratic functions of the input. We conclude that suitably crafted analog reservoirs are thus capable of matching and even surpassing the performance of standard reservoirs. The generalization error of the pure memristor network exceeded 1, and shows that the $\textrm{MSE}$ on this interval was twice the variation in $z$.  This indicates no ability to generalize the fit from the training interval and highlights the importance of reporting generalization error, rather than training error, in work on electronic reservoirs.

\begin{table}
\centering
\caption{Comparison of reservoir performance on the quadratic filtering task described in the main text.}
\begin{tabular}{l|c|l|l} \label{tab:comp}
Reservoir & dim($\vec{x}$) & $\,\nMSE$ & Gen. $\nMSE$ \\
\hline
Pure Memristor Network & 800 & \quad 0.1 &\qquad  2. \\
ESN & 780 & \quad0.02 & \qquad0.03\\
LRC$\to$Memristor & 780 & \quad0.01 & \qquad 0.01\\
\hline
\end{tabular}
\end{table}

\section*{Discussion}

Despite wide interest in utilizing electronic circuits with memory  for hardware reservoirs, an understanding of how these systems process and store information has been lacking. For echo state networks, the balance between memory and nonlinearity is controlled primarily by the spectral radius of the coupling matrix. However, no similar conditions have been explored for electronic networks.

In this work we have shown that linear electronic reservoirs of LRC circuits can be constructed with optimal memory properties, having an eigenvalue spectrum known to correspond to an extensive memory (e.g. that scales proportionally to the number of components).  This may be interpreted as performing a Fourier transform of the driving signal in hardware, where the eigenvalue spectrum required can be designed appropriately for a given problem.

In memristor reservoirs, we have shown that while the system contains contributions from terms of very high order, these are moderated in strength by powers of $\chi$. It is essential that the reservoir be able to isolate desired terms to make use of them in the training process. We have shown that using paired memristors of opposite polarity gives a substantial increase in the reservoir's ability to isolate their quadratic kernels.

Combining LRC and memristor networks into deep reservoirs allows the utilization of the memory capabilities of LRC reservoirs and the nonlinear capacities of memristor reservoirs in order to achieve specific computational goals.  Utilizing an LRC network as a deep layer allows nonlinear computations performed in the surface memristor network to be stored for long times.  Similarly, using an LRC reservoir as the surface layer to drive a deep layer memristor network will calculate products of Fourier modes and give enhanced unequal-time quadratic capacities.  Most importantly, this leads to a total quadratic memory which scales extensively in the system size, such that arbitrary products of the input can be constructed by a sufficiently large reservoir.

This analysis can have substantial impacts on performance, as we show in our comparison to ESN reservoirs.  The hybrid reservoirs we present give a 10-fold improvement over the naive memristor network implementation, and perform on par with the ESN implementation.  Properly constructed electronic reservoirs should thus be capable of matching the performance of standard reservoirs but also allow the use of larger reservoirs and faster computation times.

Our approach to the analysis of the computational capacities of memristor and LRC reservoirs can be generalized to higher order kernels of the network and to other nonlinear elements.  In this sense we present a general approach to the understanding of physical reservoirs, in analogy to the methods available to tune ESNs by trading between memory storage and nonlinearity \cite{lukovsevivcius2012practical}.



\newcommand\ackcontent{The work of FC and FCS was carried out under the auspices of the NNSA of the U.S. DoE at LANL under Contract No. DE-AC52-06NA25396. FC was also financed via DOE-ER grant PRD20190195, and FCS by a CNLS Fellowship and 20190195ER. Revision of this manuscript by FCS was performed while employed at the London Institute for Mathematical Sciences.
AK was supported by grant number FQXi-RFP-IPW-1912 from the Foundational Questions Institute and Fetzer Franklin Fund, a donor advised fund of Silicon Valley Community Foundation. AK thanks  the Santa Fe Institute for helping to support this research.}

\section*{Acknowledgements}
\ackcontent

\bibliographystyle{apalike}
\bibliography{reservoir}

\end{document}